\pdfoutput=1

\documentclass[11pt]{article}

\usepackage[preprint]{acl}

\usepackage{times}
\usepackage{latexsym}

\usepackage[T1]{fontenc}

\usepackage[utf8]{inputenc}

\usepackage{microtype}

\usepackage{inconsolata}

\usepackage{graphicx}

\usepackage{color, xcolor}
\usepackage{booktabs}
\usepackage{multicol}
\usepackage{multirow, makecell, caption}
\usepackage{colortbl}
\usepackage{tikz}
\usepackage{arydshln}
\usepackage{pgfplots}
\usepackage{amsmath}
\usepackage{amssymb}

\usepackage{tabularx}
\usepackage{enumerate}

\makeatletter
\def\adl@drawiv#1#2#3{%
        \hskip.5\tabcolsep
        \xleaders#3{#2.5\@tempdimb #1{1}#2.5\@tempdimb}%
                #2\z@ plus1fil minus1fil\relax
        \hskip.5\tabcolsep}
\newcommand{\cdashlinelr}[1]{%
  \noalign{\vskip\aboverulesep
           \global\let\@dashdrawstore\adl@draw
           \global\let\adl@draw\adl@drawiv}
  \cdashline{#1}
  \noalign{\global\let\adl@draw\@dashdrawstore
           \vskip\belowrulesep}}
\makeatother

\title{Order Matters: Investigate the Position Bias in Multi-constraint Instruction Following}

\author{
    Jie Zeng\textsuperscript{1}, Qianyu He\textsuperscript{1}, Qingyu Ren\textsuperscript{1,3}, Jiaqing Liang\textsuperscript{2\textdagger}, Yanghua Xiao\textsuperscript{1\textdagger}\\\textbf{Weikang Zhou\textsuperscript{3}, Zeye Sun\textsuperscript{3}, Fei Yu\textsuperscript{3}}\\
    \\
    \textsuperscript{1}Shanghai Key Laboratory of Data Science, School of Computer Science, Fudan University \\
    \textsuperscript{2}School of Data Science, Fudan University  \textsuperscript{3}Ant Group\\
    \{jzeng23, qyhe21, qyren24\}@m.fudan.edu.cn, \{liangjiaqing, shawyh\}@fudan.edu.cn\\
}

\begin{document}
\maketitle

\begin{abstract}

Real-world instructions with multiple constraints pose a significant challenge to existing large language models (LLMs). An observation is that the LLMs exhibit dramatic performance fluctuation when disturbing the order of the incorporated constraints. Yet, none of the existing works has systematically investigated this position bias problem in the field of multi-constraint instruction following. To bridge this gap, we design a probing task where we quantitatively measure the difficulty distribution of the constraints by a novel Difficulty Distribution Index (CDDI). Through the experimental results, we find that LLMs are more performant when presented with the constraints in a ``hard-to-easy'' order. This preference can be generalized to LLMs with different architecture or different sizes of parameters. Additionally, we conduct an explanation study, providing an intuitive insight into the correlation between the LLM's attention and constraint orders. Our code and dataset are publicly available at \url{https://github.com/meowpass/PBIF}.


\begin{figure}[t] 
    \centering
        \includegraphics[width=0.5\textwidth, height=8cm]{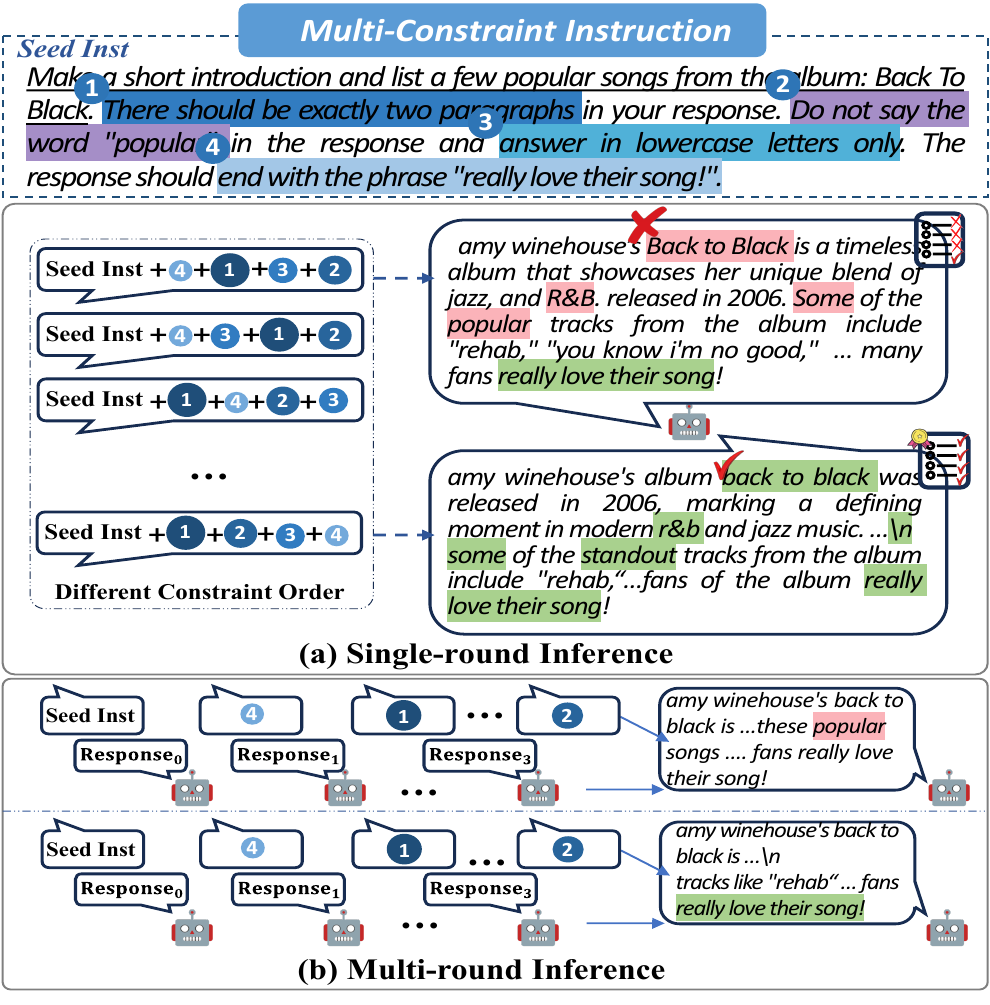}
    \caption{(a) In single-round inference, the LLM performs differently when handling the same instruction with different constraint orders. (b) In multi-round inference, the latter response is evitably affected by the former context.}
    \label{fig:intro}
\end{figure}

\end{abstract}

\section{Introduction}\label{sec:intro}


Large language models (LLMs) have made impressive progress in massive natural language tasks~\cite{wan2024tnt, zhang2024linkner} and have been applied to various real-world scenarios~\cite{bai2023qwen, bi2024deepseek}. To achieve satisfactory performance, it is crucial for LLMs to understand the user's instructions and convey desired outputs, which is known as the Instruction Following capacity of LLM~\cite{yin2023llm, xu2024wizardlm}. 

In practice, instructions are usually incorporated with multiple constraints of different types, e.g., format constraint which limits the model's output to a specific format. Nevertheless, existing LLMs often struggle to follow multi-constraint instructions, making multi-constraint instruction following an obstacle to hinder LLMs' real-world application~\cite{wen2024benchmarking, yin2023llm}.

\begin{figure*}[t] 
    \centering
            \includegraphics[width=1\textwidth]{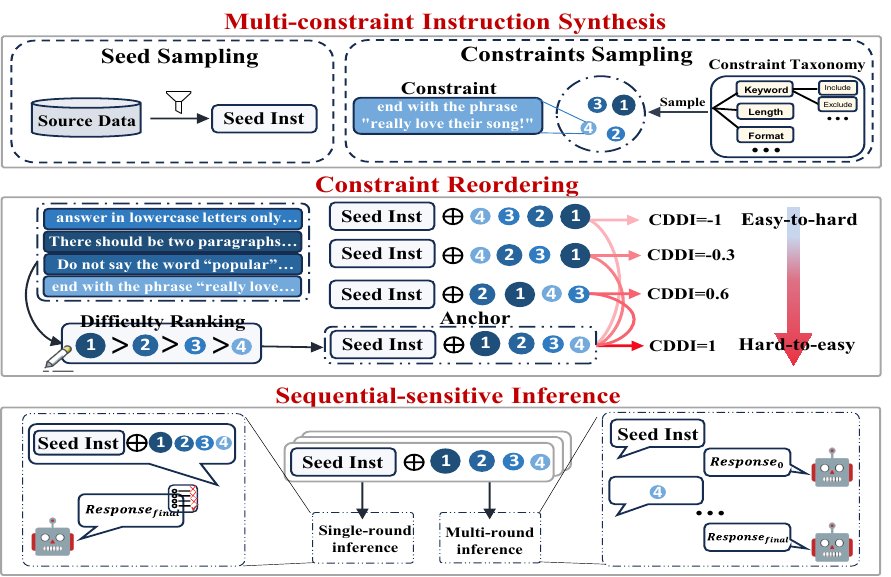}
    \caption{The procedure of the probing task. First, we synthesize the initial instructions by sampling seed instructions and corresponding constraints. Then, we obtain instructions with different constraint orders by reordering the incorporated constraints. Finally, we conduct model inference on single and multi-round settings.}
    \label{fig:method}
\end{figure*}

Recently, a lot of works have demonstrated that LLMs are sensitive to the position of the referred context in many tasks, such as multi-document question answering, text evaluation, and list-wise ranking~\cite{liu2024lost,zheng2023judging, tang2024found}. Since there are usually multiple constraints coexisting in the complex instruction, the position bias problem is also significant in multi-constraint instructions. As shown in Fig.~\ref{fig:intro}, in the single-round scenario, the LLM's performance varies significantly when presented with instructions that have different constraint orders, even though the two instructions are semantically identical. When it comes to the multi-round scenario, different constraint orders impose different impacts on the intermediate responses, thus inevitably leading to a discrepancy in the quality of the final responses.

Nevertheless, the position bias of constraint orders in the multi-constraint instruction following remains an under-explored problem. Existing work manually assigns difficulty to different constraints based on a predefined rule and orders the constraints according to their difficulty. They empirically demonstrate the existence of LLMs' performance fluctuation brought by different constraint order~\cite{chen2024sifo}. However, on the one hand, handcraft difficulty categorization fails to reflect the real difficulty disparity of different constraints~\cite{dentella2024testing, srivastava2023beyond}. On the other hand, they merely analyze the constraint order in a qualitative way, lacking a quantitative metric to measure the disparity of constraint order. Additionally, none of the existing works has provided an intuitive explanation for the position bias in multi-constraint instructions. It remains unclear how the LLMs handle instructions with different constraint orders.

To address all the problems above, we systematically investigate the position bias problem in the multi-constraint instructions. First, we propose a novel metric called the Constraint Difficulty Distribution Index (CDDI) to quantitatively describe the disparity of constraint order from the perspective of constraint difficulty. We leverage the accuracy of the LLM to quantify the difficulty of different constraints, thus precisely reflecting their disparity. Then, for a thorough study of the position bias problem, we design a probing task. As shown in Fig.~\ref{fig:method}, we construct a large number of multi-constraint instances with different constraint orders and explore two practical scenarios: single-round inference and multi-round inference. Our experiments find existing LLMs commonly perform better with the ``hard-to-easy'' constraint orders, i.e., possibly placing harder constraints in former positions. Finally, to make an intuitive explanation of our findings, we resort to a gradient-based method~\cite{wu2023language}. We visualize the importance of different constraints located in different positions. We observe that the constraint order will affect how the LLM handle the constraints and is highly correlated to the LLM's performance on a specific constraint.

In summary, our main contributions are as follows: (1) We are the first to systematically investigate the position bias problem in multi-constraint instruction following. (2) We propose a novel CDDI metric to quantify the disparity of different constraint orders in the multi-constraint instructions. (3) Through extensive experiments, we find that existing LLMs can achieve a better performance when presented with constraints in ``hard-to-easy'' orders. This finding can be generalized in both single-round and multi-round scenarios, regardless of the architecture of LLM, the size of LLM's parameters and the number of constraints. (4) Our explanation study explores how the LLMs assign attention when provided with instructions in different constraint orders and demonstrates the significant correlation between the attention patterns and the LLMs' performance on specific constraints.


\section{Related Work}

\subsection{Complex Instruction Following}
Riding on the wave of the large language model, the instruction following has attracted increasing attention for it is easy to be perceived by the users~\cite{zhou2023instructionfollowing, lou2024large}. Practical instructions are complex, usually incorporated with multiple constraints of different types~\cite{zhou2023instruction, he2024can}. A lot of evaluation benchmarks have found that multi-constraint instruction following is nontrivial for the LLMs~\cite{jiang2023followbench, wen2024benchmarking, qin2024infobench}. Consequently, several works propose to improve the LLM's complex instruction following capacity by introducing additional instruction fine-tuning~\cite{sun2024conifer, cheng2024spar, zhang2024divideverifyrefine}. 

Different from these works, we focus on the inference stage of the LLMs instead of model training. Especially, we aim to investigate the position bias problem brought by the constraint order, which poses an essential impact on the model performance.

\subsection{Position Bias in the LLM}
The position bias problem is common in the various LLM tasks~\cite{liu2024lost, zheng2023judging, zeng2023evaluating}. Researchers fisrt find that the LLM's performance degrades dramatically by merely changing the order of relevant information in the long-context question answering. A lot of works have studied the position bias problem in the field of logical reasoning~\cite{chenpremise, liu2023concise, berglund2023reversal}. They find the LLM is sensitive to the order of premises, although such ordering actually does not alter the reasoning task~\cite{chenpremise, liu2023concise}.  

Despite so, none of these works has studied the position bias problem in the field of instruction following, especially multi-constraint instruction following. SIFo~\cite{chen2024sifo} is the most related work to ours. They manually differentiate the constraints based on the context length they will influence and conduct an empirical study to verify whether the model performance will be affected by the constraint order. However, Their investigation of position bias is fairly qualitative. Different from them, we are the first to make a systematical and thorough investigation on the position bias of constraints in multi-constraint instruction following.

\section{Method}

\subsection{Background}
In this paper, we mainly focus on the multi-constraint instruction $I_c$. It can be formulated as a seed instruction incorporated with ${n}$ constraints:
\begin{equation}
\label{eq1}
    I_c = I_s \oplus C_1 \oplus ... \oplus C_n,
\end{equation}
where the seed instructions $I_s$ describe a task, e.g., write a story, while these constraints $\sum_{i=1}^n C_i$ limit the output from different aspects, e.g., format, length, content, etc. $\oplus$ stands for the concatenation operation. 

\subsection{Probing Task} \label{method}
\subsubsection{Task Formulation}
To investigate the impact of constraint order, we introduce a probing task. In this task, the LLM is given multi-constraint instructions with constraints arranged in various orders. The LLM's task is to generate a response that follows all constraints. We evaluate the LLM in two practical scenarios: single-round and multi-round inference. The LLM's responses are then evaluated to determine its performance across various constraints. The overall procedure is illustrated in Fig.~\ref{fig:method}. In the following sections, we will provide a detailed explanation.

\subsubsection{Multi-constraint Instruction Synthesis}\label{sec:ins_cons}
To ensure the generalizability of probing data, we construct the initial multi-constraint instructions which include a variety of tasks and diverse constraint combinations. The multi-constraint instruction synthesis can be further divided into two parts: seed sampling and constraint sampling. 

For the seed sampling, we sample data from three source datasets: (1) Natural Instructions V2~\cite{wang2022supernaturalinstructions}. It is an instruction collection covering more than 1600 NLP tasks. We filter those tasks that are too easy and could potentially conflict with complex constraints, e.g., object classification and sentiment tagging. Then, we randomly sample 52 instructions from the remaining tasks. (2) Self-Instruct~\cite{wang2023self}. We only sample 83 instances from their initial 175 seed instructions which are formulated by humans. (3) Open Assistant~\cite{kopf2024openassistant}. Following the strategy of Suri~\cite{li2023self}, we filter out the first turn of the conversation with the highest quality (marked as rank 0 in the dataset) and sample 65 instances from them. Overall, we obtain 200 seed instructions, where the number of instructions is denoted as $n_{seed}$.

As for the constraint sampling, we first categorize the constraints into 8 groups with 25 fine-grained types~\cite{zhou2023instructionfollowing}. For each type of constraint, we employ 8 different expressions to describe it\footnote{More details are shown in Appx.~\ref{appx:cons_tax}}. Then, we sample $n$ constraints from the constraint taxonomy and use the predefined rules to avoid possible conflicts. To ensure diversity, we repeat the sampling process to obtain $n_{cc}$ distinct constraint combinations, deriving $n_{seed}\times n_{cc}$ multi-constraint instructions.

\subsubsection{Constraint Reordering} \label{reorder}
To quantitatively construct instructions with different constraint orders, here are two questions that need to be answered: (1) \textit{How do we distinguish the disparity of different constraints}? (2) After we order the constraints based on their disparity, \textit{how do we quantitatively describe the disparity of constraint orders}?

An appropriate solution for the first question is to categorize the constraints based on their difficulty~\cite{chen2024sifo}. In this paper, we also sort the constraints based on their difficulty. However, different from existing works which designate the difficulty of the constraints based on handcraft rules, we measure the difficulty of a constraint via the overall accuracy of following it in our probing datasets. The formulation is as follows:
\begin{equation}
\label{eq2}
    \text{Dff}_{C_x}= \text{Softmax}(1-\text{Acc}_{C_{x}}), 
\end{equation}
\begin{equation}
    \label{eq3}
    \text{Acc}_{C_x} = \frac{1}{N_{x}}\sum_{i=1}^{N_{x}}c_x^i.
\end{equation}
The $C_x$ refers to a specific type of constraint, the $N_{x}$ stands for the total number of instructions corresponding to the constraint $C_{x}$, and the $c_x^i$ is a binary value to reflect whether the constraint $C_{x}$ is followed in the $i^{th}$ instruction. 

To quantitatively describe the disparity of constraint order, we propose a novel metric called the Constraint Difficulty Distribution Index (CDDI) which quantifies a specific constraint order based on its difficulty distribution. Given the difficulty of different types of constraints, we can readily attain the difficulty distribution of the constraints incorporated in the multi-constraint instructions. Specifically, for a multi-constraint instruction, we rank the incorporated constraints based on their difficulty, from the hardest to the easiest. We set this “hard-to-easy” constraint order as an anchor since it depicts an extreme situation, i.e., we designate the $\text{CDDI}=1$ when the constraints fall in this order. Consequently, akin to the Kendall tau distance~\cite{cicirello2020kendall}, we measure the difficulty distribution of a specific constraint order $o$ by comparing it with the ``hard-to-easy'' constraint order $o_{max}$. The formula is shown as:
\begin{equation}
    \label{eq4}
    \text{CDDI}_{o} = \frac{N_{con}-N_{dis}}{N_{pair}} = \frac{2(N_{con}-N_{dis})}{n(n-1)}.
\end{equation}
where $N_{con}$ and $N_{dis}$ represent the number of concordant and discordant distribution pairs of constraints between $o$ and $o_{max}$, respectively. The $N_{pair}$ is the total number of compared constraint pairs. Overall, we select $n_{dd}$ different difficulty distributions, finally comprising $n_{seed}\times n_{cc}\times n_{dd}$ instances.

\subsubsection{Sequential-Sensitive Inference}
Given the multi-constraint instructions with different constraint orders, we evaluate the model's performance in two common scenarios: single-round inference and multi-round inference. In single-round inference, the LLM is directly given the multi-constraint instructions with different constraint distributions. We argue that different constraint distributions could impose different levels of difficulty on the LLM to handle. The multi-round inference introduces a more typical setting: the user will first provide the LLM with the core intention (i.e., the seed instruction in this work), and then iteratively put forward the constraints in order to obtain a final response.

To evaluate the model performance, apart from the constraint following accuracy mentioned in Eq.(\ref{eq3}), we also verify its constraint-level accuracy $Acc_{cons}$ and instruction-level accuracy $Acc_{inst}$. Corresponding formulas are shown below:
\begin{equation}
    \label{eq5}
    \small
    \text{Acc}_{\text{cons}} = \frac{1}{mn}\sum_{i=1}^{m}\sum_{j=1}^{n}c_i^j,     \text{Acc}_{\text{inst}} = \frac{1}{m}\sum_{i=1}^{m}\prod_{j=1}^{n}c_i^j.
\end{equation}
where $m$ and $n$ refer to the number of instructions and constraints in the instruction, respectively. Similar to Eq.(\ref{eq3}), the $c_i^j$ is a binary value which equals 1 when the constraint is followed in the $i^{th}$ instruction. All the evaluation is conducted by leveraging the script introduced in ~\cite{zhou2023instructionfollowing}. We only evaluate the final responses produced by the LLMs.

\begin{figure}[t] 
    \centering
        \includegraphics[width=0.5\textwidth]{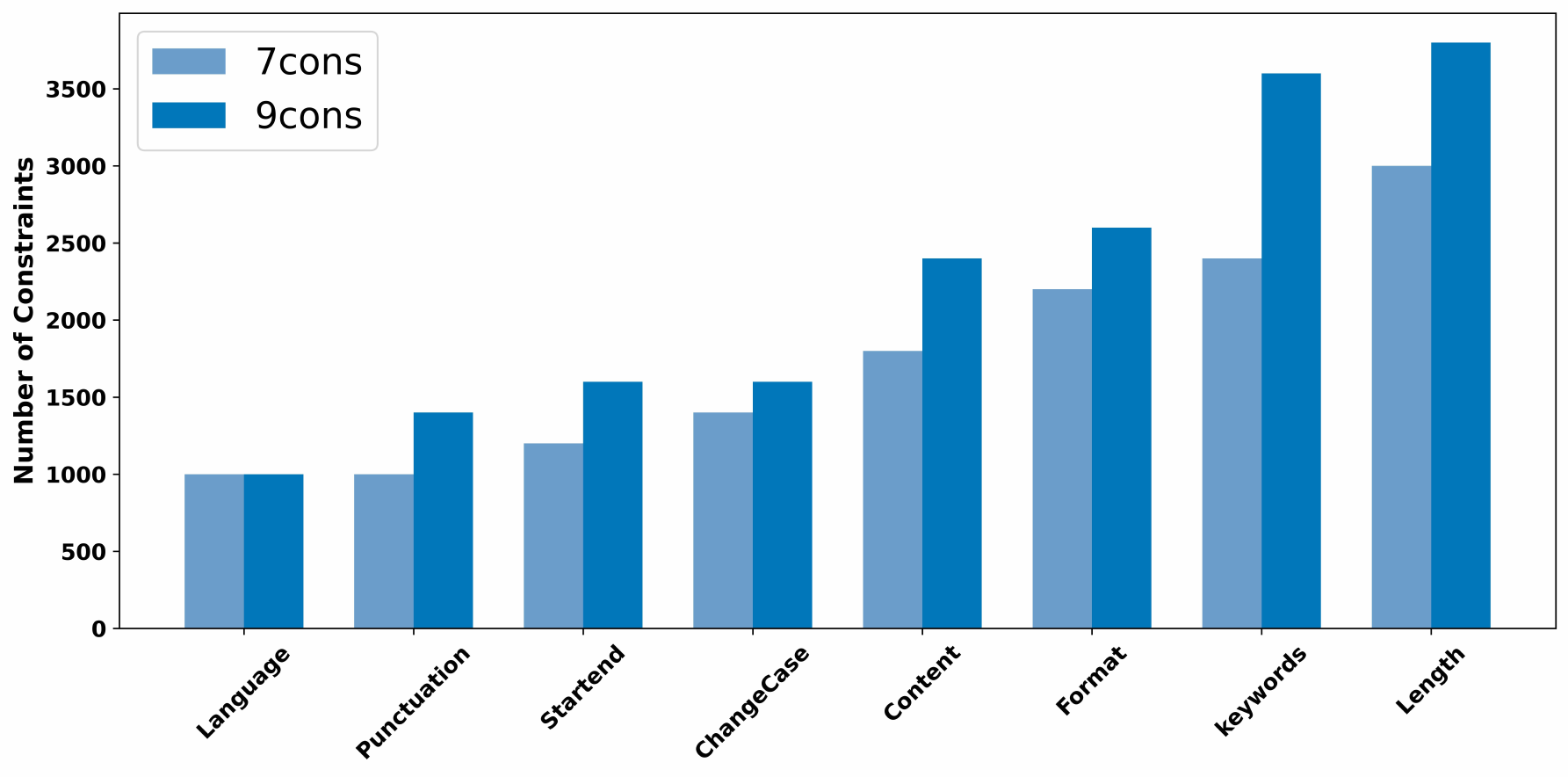}
    \caption{The statistic of different types of constraints in the probing data. The 7cons and 9cons stand for the setting when $n$=7 and $n$=9, respectively.}
    \label{fig:statistic}
\end{figure}

\section{Empirical Study}

\begin{figure*}[t] 
    \centering
        \includegraphics[width=0.9\textwidth, height=5cm]{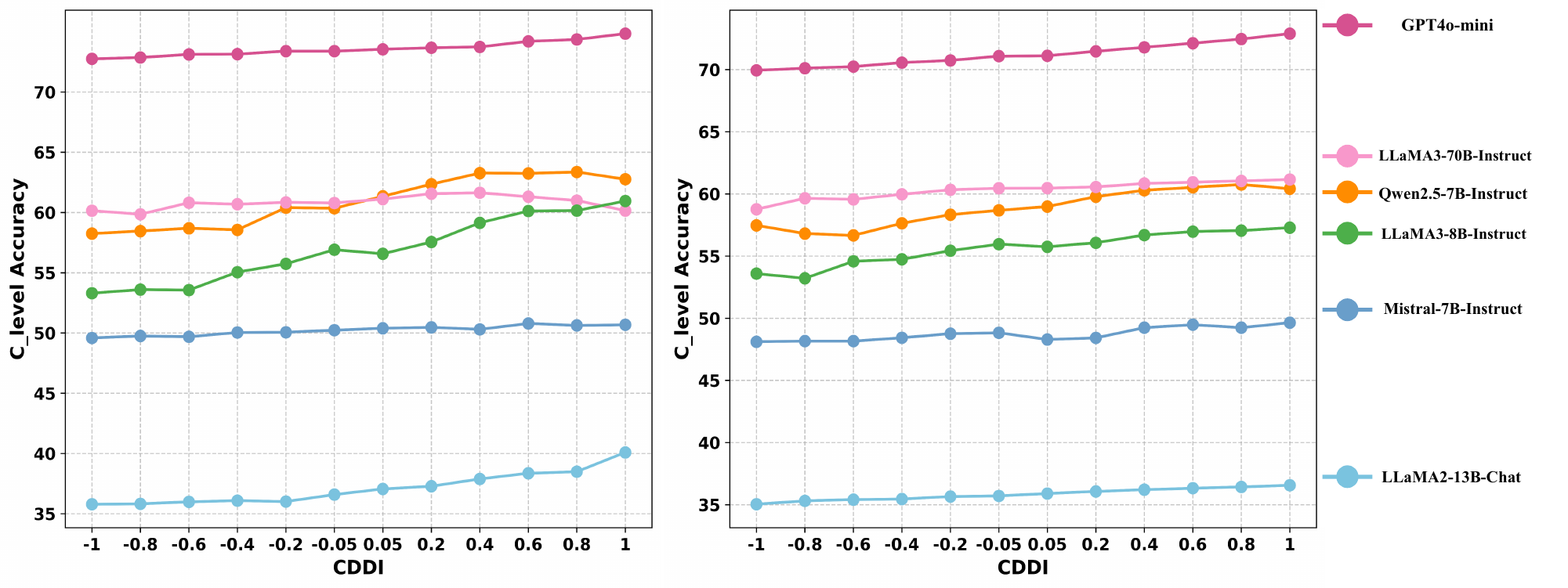}
    \caption{The performance of different LLMs in the single-round inference. The left and right figures show the results with the number of constraints $n$ set to 7 and 9, respectively. With the increase of the CDDI, the constraint order changes from ``easy-to-hard'' to ``hard-to-easy''.}
    \label{fig:7cons}
\end{figure*}

\begin{figure*}[t] 
    \centering
        \includegraphics[width=0.9\textwidth,height=5cm]{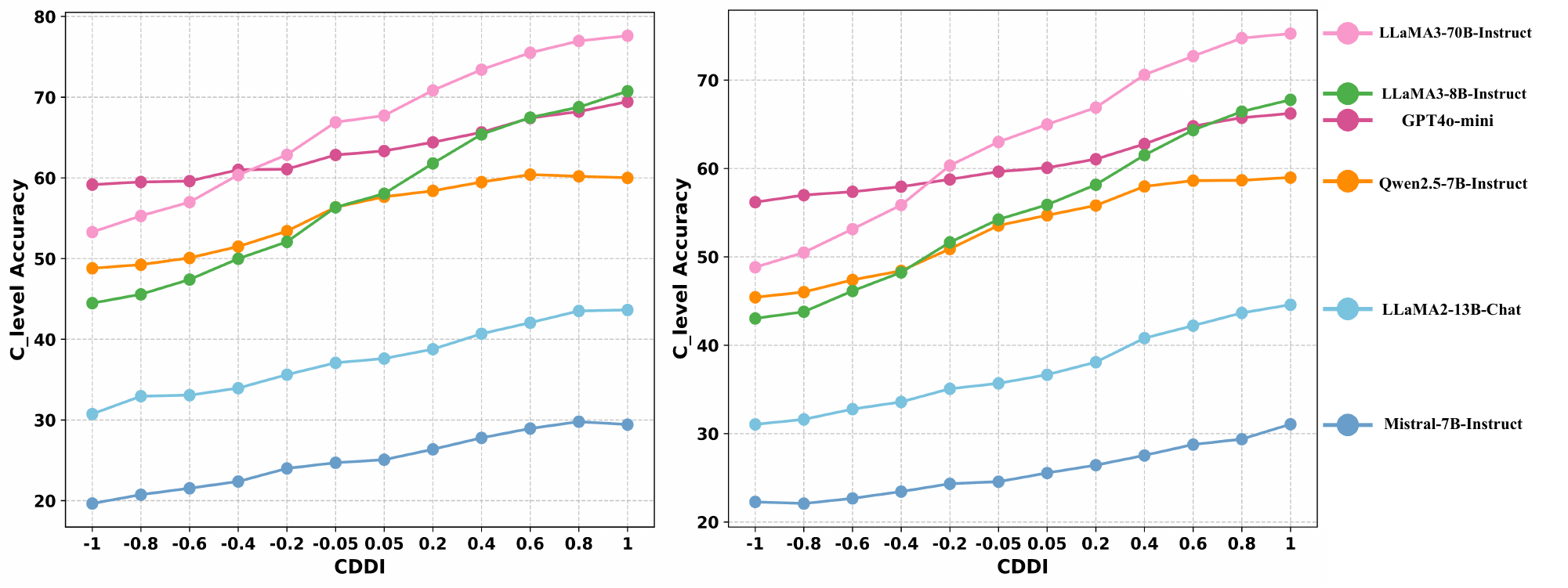}
    \caption{The performance of different LLMs in the multi-round inference. The left and right figures show the results with the number of constraints $n$ set to 7 and 9, respectively. With the increase of the CDDI, the constraint order changes from ``easy-to-hard'' to ``hard-to-easy''.}
    \label{fig:multi_7cons}
\end{figure*}

\begin{table*}[t]
\newcolumntype{g}{>{\columncolor{green!4}}r}
\newcolumntype{b}{>{\columncolor{blue!4}}r}
\renewcommand{\arraystretch}{0.9} 
\renewcommand{\familydefault}{\rmdefault}
\resizebox{\textwidth}{!}{
\begin{tabular}{cccccccccgb}
\toprule
\textbf{CDDI} & \textbf{Length} & \textbf{Language} & \textbf{Punctuation} & \textbf{Format} & \textbf{Keywords} & \textbf{ChangeCase} & \textbf{Startend} & \textbf{Content} & \textbf{C\_level} & \textbf{I\_level} \\ \midrule
\rowcolor[gray]{0.95}\multicolumn{11}{c}{\textit{Single-round Inference}} \\ 
\textbf{-1}  & 27.50 & 28.20  & 23.30 & 71.14 & 68.58 & 49.57 & \underline{62.92} & \textbf{81.22} & 53.30 & 1.95  \\ 
\textbf{-0.8} & 28.23 & 30.70  & 23.90 & 73.64 & 68.46 & 49.00 & \textbf{63.50} & 77.78 & 53.60 & 1.80  \\ 
\textbf{-0.6} & 28.53 & 31.10  & 26.60 & 71.23 & 69.25 & 49.79 & 60.92 & \underline{78.22} & 53.56 & 1.95 \\ 
\textbf{-0.4} & 28.53 & 36.10  & 30.70 & 72.41 & 71.58 & 51.64 & 62.33 & 77.83 & 55.05 & 2.10  \\ 
\textbf{-0.2} & 29.33 & 39.30  & 35.30 & 73.82 & 72.08 & 50.07 & 60.75 & 77.44 & 55.74 & 2.40  \\ 
\textbf{-0.05} & \textbf{30.27} & 42.90 & 36.80 & 74.95 & 73.46 & 52.14 & 60.50 & 77.50 & 56.91 & 2.90  \\ \cdashlinelr{1-11} 
\textbf{0.05}  & 29.17 & 46.70  & 38.00 & 72.68 & 74.75 & 50.79 & 61.50 & 75.33 & 56.57 & 2.75  \\ 
\textbf{0.2}   & 28.17 & 50.50  & 43.30 & 76.05 & 75.92 & 52.07 & 61.42 & 72.94 & 57.55 & 2.75  \\ 
\textbf{0.4}   & 30.23 & 54.50  & 46.40 & 76.64 & 76.29 & 54.07 & 62.83 & 74.17 & 59.14 & 2.65  \\ 
\textbf{0.6}   & 29.83 & 59.20  & 49.70 & \textbf{79.09} & \underline{77.42} & 56.71 & 58.58 & 74.33 & 60.12 & 3.00  \\ 
\textbf{0.8}   & 29.40 & \underline{60.50}  & \underline{51.70} & 77.91 & \textbf{77.63} & \underline{58.07} & 58.25 & 73.89 & \underline{60.16} & \underline{3.05}  \\ 
\textbf{1}     & \underline{30.03} & \textbf{67.10} & \textbf{53.10} & \underline{78.00} & 77.21 & \textbf{59.21} & 57.42 & 74.61 & \textbf{60.95} & \textbf{3.50}  \\
\rowcolor[gray]{0.95}\multicolumn{11}{c}{\textit{Multi-round Inference}} \\ 
\textbf{-1}  & \textbf{62.60} & 64.00 & 54.20 & 21.59 & 57.79 & 62.50 & 10.83 & 16.61 & 44.47 & 0.75  \\ 
\textbf{-0.8} & \underline{59.63} & 64.90 & 61.50 & 22.27 & 62.46 & 61.93 & 13.17 & 17.39 & 45.57 & 0.75  \\ 
\textbf{-0.6} & 54.65 & 67.87 & 65.67 & 25.74 & 67.83 & 59.47 & 22.75 & 20.08 & 47.40 & 0.65  \\ 
\textbf{-0.4} & 52.77 & 68.74 & 64.46 & 32.30 & 69.57 & 61.44 & 30.78 & 26.21 & 49.98 & 1.05  \\ 
\textbf{-0.2} & 48.73 & 68.74 & 62.42 & 38.67 & 74.67 & 59.16 & 39.00 & 32.02 & 52.07 & 1.25  \\ 
\textbf{-0.05} & 46.48 & 69.97 & 67.17 & 46.38 & 76.04 & 60.97 & 48.79 & 46.58 & 56.35 & 1.40  \\ \cdashlinelr{1-11} 
\textbf{0.05}  & 45.32 & 70.08 & 68.84 & 51.19 & 76.62 & 62.18 & 52.68 & 50.47 & 58.04 & 1.80  \\ 
\textbf{0.2}   & 44.81 & 69.91 & 66.73 & 58.35 & 80.20 & 60.34 & 62.41 & 63.22 & 61.79 & 3.41  \\ 
\textbf{0.4}   & 44.30 & \textbf{72.50} & \underline{69.10} & 64.50 & 81.75 & \underline{63.50} & 68.00 & 73.56 & 65.39 & \underline{5.60}  \\ 
\textbf{0.6}   & 43.71 & 68.87 & 68.20 & 71.71 & 83.87 & 59.83 & 71.98 & 81.77 & 67.47 & 5.05  \\ 
\textbf{0.8}   & 44.35 & 68.37 & 68.00 & \underline{75.94} & \underline{84.49} & 61.54 & \underline{70.31} & \underline{84.88} & \underline{68.76} & \textbf{6.00}  \\ 
\textbf{1}     & 44.07 & \underline{70.90} & \textbf{69.60} & \textbf{81.41} & \textbf{85.08} & \textbf{64.43} & \textbf{72.58} & \textbf{87.22} & \textbf{70.74} & 4.00  \\
\bottomrule
\end{tabular}
}
  \caption{
  The overall performance of LLaMA3-8B-Instruct on multi-constraint instructions with different CDDI values. From left to right, we sort the constraint types from the hardest to the easiest.
  }
  \label{tab:main}
\end{table*}

\subsection{Experiment Setup}
\paragraph*{Models} For our probing task, to ensure the generalizability of our study, we conduct experiments on both closed and open-source LLMs with varying architectures and parameter sizes. Specifically, we introduce the following models: (1) LLaMA3-8B-Instruct and LLaMA3-70B-Instruct~\cite{dubey2024llama}. (2) LLaMA2-13B-Chat~\cite{touvron2023llama}. (3) Mistral-7B-Instruct~\cite{jiang2023mistral}.\footnote{We use the latest v0.3 version.} (4) Qwen2.5-7B-Instruct~\cite{yang2024qwen2}. (5) GPT4o-mini~\cite{achiam2023gpt}.


\paragraph*{Datasets} We construct various multi-constraint instructions with different constraint orders (Sec.\ref{method}). We empirically set the number of constraints ${n}$ to 7. To ensure the diversity and complexity, we set the number of constraint combinations $n_{cc}$ to 10 and the number of difficulty distributions $n_{dd}$ to 12, finally obtaining $200\times10\times12=24\text{K}$ samples. To verify the influence of constraint number, we also conduct experiments on the setting when ${n=9}$. The statistic of the data for the probing task is provided in Fig.~\ref{fig:statistic}.

\subsection{Results}
\paragraph*{LLMs prefer to “hard-to-easy” constraint distribution.} As shown in Fig.~\ref{fig:7cons}, most of the LLMs exhibit a dramatic performance fluctuation on instructions with varying constraint distributions. When the constraint number is set to 7, the LLaMA3-8B-Instruct and Qwen2.5-7B-Instruct show approximately 7$\%$ and 5$\%$ performance disparity in extreme situations. This indicates the vulnerability of existing LLMs to the position bias brought by the constraint order. Also, the LLMs tend to be more performant to instructions with higher CDDI values. Even the LLaMA3-70B-Instruct exposes a clear preference for higher CDDI value as the number of constraints increases to 9, demonstrating that ``hard-to-easy'' is a superior constraint distribution for existing LLMs.

\paragraph*{Multi-round inference exhibits more severe position bias compared with the single-round inference.} The LLMs' performance in multi-round inference is presented in the Fig.~\ref{fig:multi_7cons}. Compared with the results in the single-round inference, the performance gap becomes more prominent. All the LLMs gain approximately 10$\%$ improvement on C$\_$level accuracy. Surprisingly, the  LLaMA3-8B-Instruct and LLaMA3-70B-Instruct achieve approximately 25$\%$ performance improvement by changing the constraint distribution from ``easy-to-hard'' (CDDI=-1) to ``hard-to-easy'' (CDDI=1). This indicates that the LLMs are more sensitive to the position bias problem in a multi-round scenario.

\paragraph*{LLMs perform better in multi-round inference when provided with the instructions in appropriate constraint order} Comparing the results in single-round (Fig.~\ref{fig:7cons}) and multi-round inference (Fig.~\ref{fig:multi_7cons}), we observe that the LLMs reach better performance if the incorporated constraints are arranged in an appropriate order. Specifically, when the CDDI value is negative, the performance of LLMs in multi-round inference lags behind that in single-round inference. Nevertheless, with the increase of the CDDI value, the LLMs can achieve superior performance in multi-round inference and reach their best performance in CDDI=1. An exception is the Mistral-7B-Instruct-v0.3. We attribute this to its inferiority in processing multi-round information~\cite{chen2024sifo}.

\paragraph*{Position bias varies in different types of constraints.} We present the performance of the LLaMA3-8B-Instruct across different types of constraints in Tab.~\ref{tab:main}. As observed, with the increase of the CDDI value, the model's performance across most constraint types shows an upward trend except for Startend and Content, indicating that not all the constraints can benefit from the ``hard-to-easy'' constraint distribution in single-round inference. We make a more comprehensive explanation study in Sec.~\ref{sec:experiment2} for further investigation. Regarding the multi-round inference, the model's performance only exhibits a drop tendency in the Length type as the CDDI value increases, indicating that the LLMs struggle to generate a length-controlled final response when the length constraint is applied early in the multi-round inference~\cite{yuan2024following}.


\begin{table*}[t]
\newcolumntype{g}{>{\columncolor{green!10}}c}

\newcolumntype{b}{>{\columncolor{blue!10}}c}
\renewcommand{\arraystretch}{0.8} 
\renewcommand{\familydefault}{\rmdefault}
\resizebox{\textwidth}{!}{
\begin{tabular}{ccccccccccc}
\toprule
\textbf{Round} & \textbf{Length} & \textbf{keywords} & \textbf{language} & \textbf{ChangeCase} & \textbf{Format} & \textbf{Content} & \textbf{Startend} & \textbf{Punctuation} & \textbf{C\_level} & \textbf{I\_level} \\ 
\midrule
\textbf{1} & 29.93 & 73.46 & 44.40 & 50.68 & 76.59 & 77.11 & 59.92 & 34.40 & 56.01 & 2.70 \\
\textbf{2} & 29.83 & 73.29 & 43.80 & 50.79 & 73.36 & 78.17 & 61.50 & 32.60 & 55.49 & 2.65 \\
\textbf{3} & 30.27 & 73.46 & 42.90 & 52.14 & 74.95 & 77.50 & 60.50 & 36.80 & 56.91 & 2.90 \\
\cdashlinelr{1-11} 
 & 30.01$\pm$0.23  & 73.40$\pm$0.10  & 43.70$\pm$0.75 & 51.20$\pm$0.82 & 74.97$\pm$1.61 & 77.59$\pm$0.53  & 60.64$\pm$0.80 & 34.60$\pm$2.11 & 56.14$\pm$0.72 & 2.75$\pm$0.13 
 \\
\bottomrule
\end{tabular}
}
  \caption{
  The performance of LLaMA3-8B-Instruct when given the multi-constraint instruction in different constraint orders while sharing the same CDDI value. By calculation, we obtain the P-value=0.9979. 
  }
  \label{tab:sensitivity}
\end{table*}

\subsection{Robustness of CDDI}
Since the CDDI is calculated by comparing the concordant and discordant pairs of two different constraint orders, there are usually multiple constraint orders sharing the same CDDI value. Therefore, we conduct a testing experiment to assess whether the LLM exhibits significant fluctuations across different constraint orders with the same CDDI value. Specifically, we set the CDDI to -0.05, a value that includes the most constraint orders in our setting, and conduct single-round inference for 3 times. The experiment results are shown in Tab~\ref{tab:sensitivity}. We calculate the P-value of the data, finding that the P-value is much larger than 0.05. This indicates that the fluctuation of LLM's performance is negligible among different constraint orders in the same CDDI value.

\section{Explanation Study}

\begin{figure*}[t] 
    \centering
        \includegraphics[width=1\textwidth]{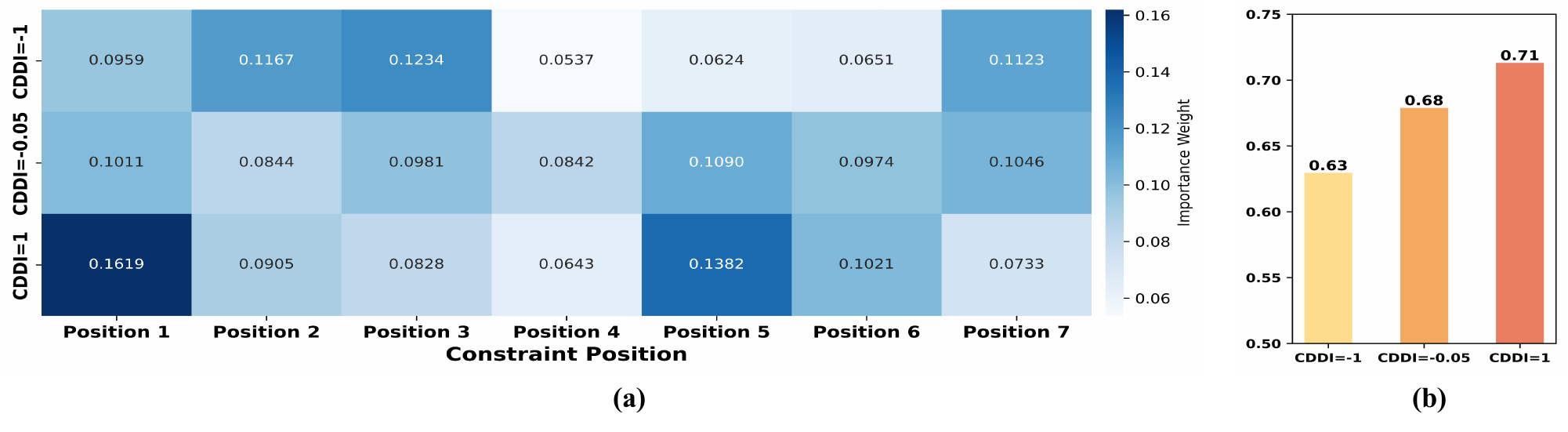}
    \caption{(a) The importance weights assigned by the LLM when handling constraints in different positions. (b) The total importance weights which designated to the constraint part in the multi-constraint instructions among three different constraint distributions.}
    \label{fig:position_score}
\end{figure*}

\begin{figure}[t] 
    \centering
        \includegraphics[width=0.48\textwidth]{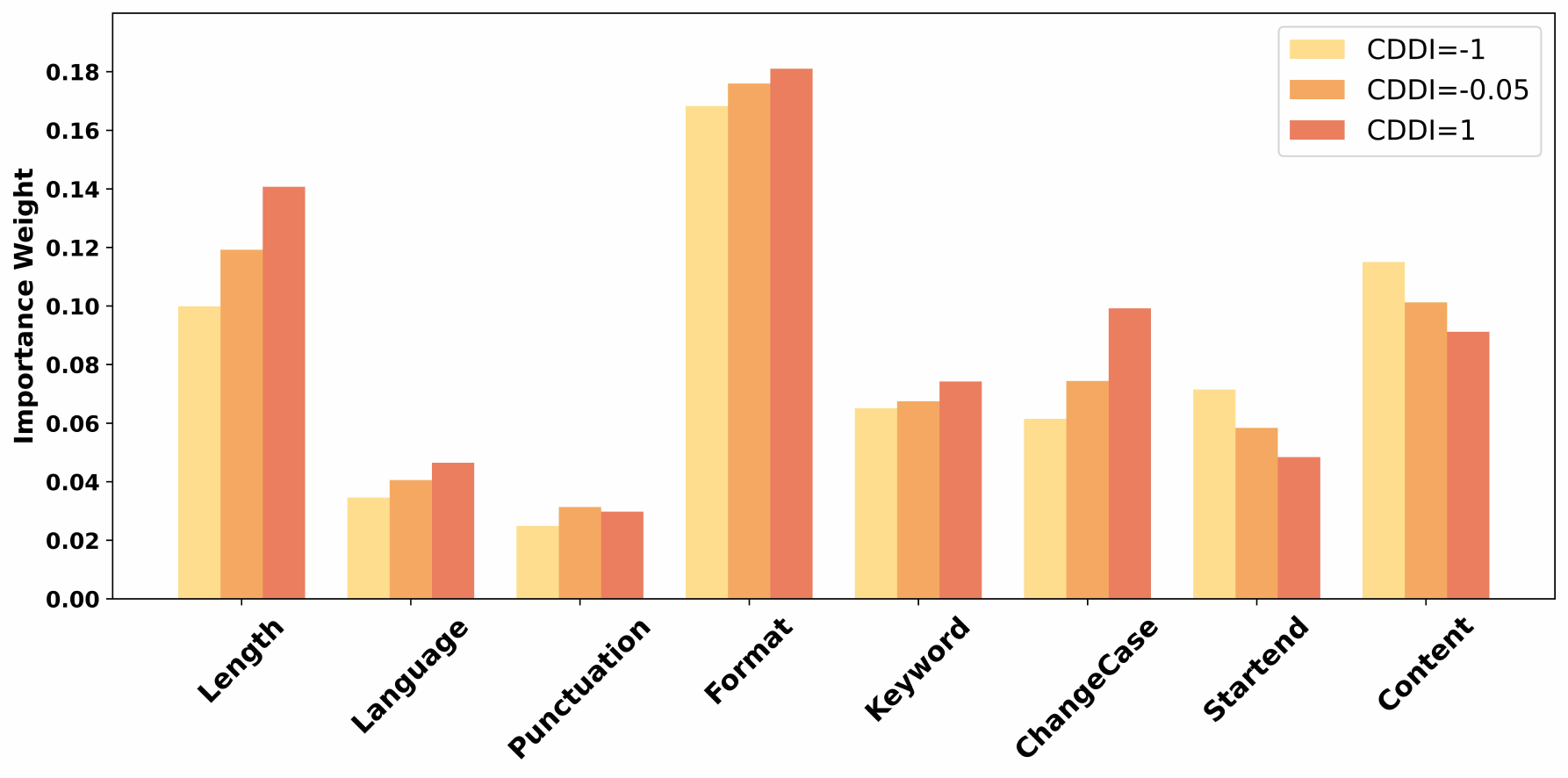}
    \caption{The importance weights across different types of constraint in three different constraint distributions.}
    \label{fig:type_score}
\end{figure}

\subsection{Explanation Metric}
To make an explanation for the influence brought by the constraints of different orders, we make an explanation study on where the LLMs mainly focus when handling multi-constraint instructions via a feature attribution-based explanation method~\cite{li2016visualizing, wu2020perturbed}. Specifically, we leverage the importance of the input tokens to measure the LLMs' attention to them. To obtain the importance of a specific instruction token $t_x$ to a response token $t_y$, we calculate the confidence change after the removal of the $t_x$, as formulated below:
\begin{equation}
    \label{eq6}
    I_{t_x,t_y}=p(t_y|Z_y)-p(t_y|Z_{y,/t_x}),
\end{equation}
where $p(\cdot|\cdot)$ is the conditional probability produced by the LLM $f$, $Z_y$ is the tokens before the $t_y$ and $Z_{y,/t_x}$ is the tokens of $Z_y$ after removing the token $t_x$. To reduce the computation, we approximate the $I_{t_x,t_y}$ with the first-order gradient $\frac{\partial f\left(t_y \mid Z_y\right)}{\partial \mathbf{E}\left[t_x\right]}$ ~\cite{wu2023language}, where $\mathbf{E}\left[t_x\right]$ is the token embedding of $t_x$. We normalize the importance $I_{t_x,t_y}$ and obtain the standard importance $S_{t_x,t_y}$ with the formula:
\begin{equation}
    \label{eq7}
    S_{t_x,t_y}= \frac{L\times I_{t_x,t_y}}{{\max_{i=1}^{N_{X}}}I_{t_i,t_y}},
\end{equation}
where $N_X$ is the number of instruction tokens and $L$ is a hyper-parameter which helps to filter the noise brought by the first-order approximation. To visualize the LLMs' attention to different constraints, we calculate the importance weight of a specific constraint $C_x$ to the final response $Y$ with the formula:
\begin{equation}
    \label{eq8}
    S_{C_x,Y}=\frac{1}{N_Y}\sum_{t_y\in Y}\sum_{t_x\in C_x}S_{t_x,t_y},
\end{equation}
where $N_Y$ is the number of response tokens.

\subsection{Experiment Set-up}
We conduct our explanation study on the LLaMA3-8B-Instruct model. We set the hyper-parameter $L$ to 10 in Eq.(\ref{eq7}) and select three most typical difficulty distributions: hard-to-easy (indicated by CDDI=1), easy-to-hard (indicated by CDDI=-1) and random (indicated by CDDI=-0.05) to conduct our experiments. We randomly sample 200 instances from the corresponding data which fall in the required CDDI value in the probing task to serve as the dataset.

\subsection{Results}
\paragraph*{Hard-to-easy constraint order induces the LLM to pay more attention to the constraint part in the multi-constraint instructions.} We visualize the importance weights of the model on the constraints in different positions. As shown in Fig.~\ref{fig:position_score} (a), in the multi-constraint instruction following, the model's attention on different positions varies with changes in the constraint orders. Specifically, when the constraints are randomly distributed across different positions (represented by CDDI=-0.05), the model assigns similar attention to all positions. As the constraint order becomes more structured (represented by CDDI=-1 and CDDI=1), the model's attention neither exhibits the “lost in the middle” phenomenon observed in long-context processing~\cite{liu2024lost}, nor a simply sequential distribution, but follows an iterative, laddered order. Then, in Fig.~\ref{fig:position_score} (b), we present the total importance weight the model assigns to the constraint part. We observe that the “hard-to-easy” constraint order attracts the most attention from the model towards the constraint part, which provides an explanation for the superiority of this constraint order.

\paragraph*{The LLM's performance on various constraints is strongly correlated with its attention patterns.} The importance weights of the model on different types of constraints are presented in Fig.~\ref{fig:type_score}. Among the three distinct difficulty distributions, the “hard-to-easy” (represented by CDDI = 1) assigns the highest importance weights to various types of constraints except for the Content and Startend. It is worth noting that this is exactly in accord with quantitative results in Tab.~\ref{tab:main}, i.e., as the CDDI value increases, the model's performance on the Content and Startend constraints shows a decreasing trend instead. Overall, the results show that the model's accuracy in following a specific type of constraint is strongly correlated with the attention assigned to it by the model. \label{sec:experiment2}

\section{Conclusion}

In this paper, we systematically investigate the position bias problem in the multi-constraint instruction following. To quantitatively measure the disparity of constraint order, we propose a novel Difficulty Distribution Index (CDDI). Based on the CDDI, we design a probing task. First, we construct a large number of instructions consisting of different constraint orders. Then, we conduct experiments in two distinct scenarios. Extensive results reveal a clear preference of LLMs for ``hard-to-easy'' constraint orders. To further explore this, we conduct an explanation study. We visualize the importance of different constraints located in different positions and demonstrate the strong correlation between the model's attention distribution and its performance.

\section{Limitations}

Our work mainly focuses on the position bias problem in the multi-constraint instruction following. We make a quantitative analysis of the influence brought by different constraint orders in the instructions. However, there are still some limitations. The constraints in our work are usually parallel to each other, which means the order change will not affect the semantic meaning of the instructions. The position bias problem for for those sequential constraints need to be further explored. Moreover, we only investigate the phenomenon of position bias in existing LLM without offering a solution. In further work, we will conduct a further probing task in sequential constraints to improve the generalization of our findings.



\bibliography{acl_latex}

\appendix

\section{Appendix}

\subsection{Implementation Details}
We utilize 8 NVIDIA A800 80GB GPUs to conduct all the experiments. We employ the vLLM framework~\cite{kwon2023efficient} to accelerate the model inference. For reproducibility, we employ the greed search in the whole inference (i.e., setting the ``do$\_$sample'' to false.).

\subsection{More details for Comstraint Sampling} \label{appx:cons_tax}
In this work, We categorize the constraints into 8 different groups. The categorization is shown in the Tab.~\ref{tab:ifeval}. For each group, there are multiple types of constraints. Specifically, the constraints are designated to: (1) Keyword constraints. These constraints focus on controlling the inclusion or exclusion of specific words or phrases within the response. (2) Language constraints. Language constraints govern the linguistic properties of the response, including the language in which the response is written (e.g., English). (3) Length constraints. These constraints focus on controlling the overall length of the response, including the number of paragraphs, words, and sentences. (4) Content Constraints. Content-related constraints define additional rules to ensure the response contains specific elements. (5) Format constraints. Formatting constraints focus on how the response is structured and styled. For example. (6) ChangeCase Constraints. These constraints focus on adjusting the case of words in the response. They may require the entire response to be in uppercase letters (e.g., ALL CAPS), or entirely in lowercase letters (e.g., all lowercase). (7) StartEnd constraints. These constraints limit the very beginning or ending of the model outputs. (8) Punctuation constraints. These constraints limit the appearance of specific commas.

\renewcommand{\arraystretch}{1.15}
\begin{table*}
\centering
\begin{tabular}{>{\raggedright}m{3cm}|m{4cm}|m{9cm}}
\hline
\textbf{Constraint Group} & \textbf{Constraint} & \textbf{Description Example} \\
\hline
\multirow{4}{*}{Keyword} & Include Keywords & Include keywords \texttt{[keyword1], [keyword2]} in your response. \\
\cline{2-3}
 & Exclude Keywords & Do not include keywords \texttt{[forbidden words]} in the response. \\
\cline{2-3}
 & Keyword Frequency & In your response, the word should appear \texttt{N} times. \\
\cline{2-3}
 & Letter Frequency & In your response, the letter \texttt{[letter]} should appear \texttt{[N]} times. \\
\hline
Language & Response Language & Your ENTIRE response should be in \texttt{[language]}, no other language is allowed. \\
\hline
\multirow{4}{*}{Length} & Number Paragraphs & Your response should contain \texttt{[N]} paragraphs. You separate paragraphs using the markdown divider \texttt{***}. \\
\cline{2-3}
 & Number Words & Answer with at least/around/at most \texttt{[N]} words. \\
\cline{2-3}
& Number Sentences & Answer with at least/around/at most \texttt{[N]} sentences. \\
\cline{2-3}
& Number Paragraphs + First Word in i-th Paragraph & There should be \texttt{[N]} paragraphs. Paragraphs and only paragraphs are separated with each other by two line breaks. The \texttt{[i]}-th paragraph must start with \texttt{[first\_word]}. \\
\hline
\multirow{2}{*}{Content} & Postscript & At the end of your response, please add a postscript starting with \texttt{[postscript marker]}. \\
\cline{2-3}
 & Number Placeholder & The response must contain at least \texttt{[N]} placeholders representing the word space brackets, such as \texttt{[address]}. \\
\hline
\multirow{6}{*}{Format} & Number Bullets & Your response must contain exactly \texttt{[N]} bullet points. Use the markdown bullet points such as: * This is a pont. \\
\cline{2-3}
& Title & Your answer must contain a title, wrapped in double angular brackets, such as \texttt{<<option of joy>>}. \\
\cline{2-3}
 & Choose From & Your response should contain one of the following options: \texttt{[options]}. \\
\cline{2-3}
 & Minimum Number Highlighted Section & Highlight at least \texttt{[N]} sections in your answer with markdown, i.e.  *highlighted section*. \\
\cline{2-3}
 & Multiple Sections & Your response must have \texttt{[N]} sections. Mark the beginning of each section with \texttt{[section\_splitter]} X. \\
\cline{2-3}
 & JSON Format & Entire output should be wrapped in JSON format. \\
\hline
\multirow{3}{*}{ChangeCase} & All Uppercase & Your entire response should be in English, capital letters only. \\
\cline{2-3}
 & All Lowercase & Your response should be in English, and in all lowercase letters. No capital letters are allowed. \\
\cline{2-3}
& Frequency of All-capital Words & In your response, words with all capital letters should appear at least \texttt{[N]} times. \\
\hline
\multirow{2}{*}{StartEnd} & End Checker & Your response must finish with this phrase: \texttt{<end\_phrase>}.  \\
\cline{2-3}
& Quotation & Wrap uour entire response with double marks. \\
\hline
Punctuation & No Commas & In your entire response, refrain from the use of any commas. \\
\hline
\end{tabular}
\caption{The categorization for different constraints.}
\label{tab:ifeval}
\end{table*}

Considering the LLM is vulnerable to different descriptions of the constraints~\cite{yan2024contrastive}, we employ the GPT4o-mini to generate different descriptions of the same constraints. Specifically, given a description example, we leverage the prompt shown in the Tab.~\ref{tab:100diverse} to seven distinct variants. Overall, we obtain 8 distinct descriptions for a specific type of constraint.

\begin{table*}[t]
\small
    \begin{tabularx}{\linewidth}{X}
    \toprule
    \color{gray}{/* \textit{Task prompt} */}\\
    You are provided with a <constraint> in an instruction. As a prompt engineer, your task is to rephrase the provided <constraint> to make it more diverse. You ought to provide five more variants of the <constraint>. Make sure your revision does not change the meaning of the original <constraint>. \\
    \color{gray}{/* \textit{Example} */}\\
    ---INPUT--- \\
    <constraint>:\\
    Your response should contain at least 3 sentences.\\
    ---OUTPUT---\\
    variants:\\
    1. Respond with at least three sentences\\
    2. Use at least 3 sentences in your reply\\
    3. Your entire response should include at least three sentences\\
    4. Organize your entire response in at least 3 sentences\\
    5. Please make sure the response is at least 3 sentences long\\
    \color{gray}{/* \textit{Input} */}\\
    ---INPUT---\\
    <constraint>:\\
    \{\textbf{Given\_constraint}\}\\
    ---OUTPUT---\\
    variants:\\
    \bottomrule
    \end{tabularx}
  \caption{
    The prompts for diversifying the descriptions of a given constraint. We utilize one-shot in-context learning to enhance the performance. The information that requires manual input is highlighted.
  }
  \label{tab:100diverse}
\end{table*}

\end{document}